# Probabilistic contingent planning based on HTN for high-quality plans

Peng Zhao

**Abstract.** Deterministic planning assumes that the planning evolves along a fully predicable path, and therefore it loses the practical value in most real projections. A more realistic view is that planning ought to take into consideration partial observability beforehand and aim for a more flexible and robust solution. What's more significant, it is inevitable that the quality of plan varies dramatically in the partially observable environment. In this paper we propose a probabilistic contingent Hierarchical Task Network (HTN) planner, named High-Quality Contingent Planner (HQCP), to generate high-quality plans in the partially observable environment. The formalisms in HTN planning are extended into partial observability and are evaluated regarding the cost. Next, we explore a novel heuristic for high-quality plans and develop the integrated planning algorithm. Finally, an empirical study verifies the effectiveness and efficiency of the planner both in probabilistic contingent planning and for obtaining high-quality plans.

## 1. Introduction

Classical planning assumes the planning environment is fully observable, namely the planer has the complete knowledge regarding the state of the world [1,2]. However, since it is tough or even impossible for an agent to ascertain the complete information at run-time, the practical environment is partially observable to the planner. Therefore it motivates the agent to plan under partial observability by taking into consideration the incomplete knowledge beforehand and generating a more robust plan [3].

Furthermore, as the uncertain states in the partially observable environment is unknown before being sensed, the quality of plans varies dramatically [4]. It results in that only planning for the feasible plans cannot meet the requirement. Planning for high-quality plans presents searching for the solution with less or the least cost, which is considered as a difficult problem [5-7]. We take the fire emergency evaluation as example. There are several dwellers trapped in a burning building. They may be in some rooms but their specific locations cannot be predetermined. There are many feasible evacuation plans but they have different costs. The goal of planning in this domain is to evacuate all the dwellers with the least cost. Therefore, planning under partial observability for high-quality plans is a challenging and valuable problem.

Hierarchical task network (HTN) planning [8,9] is an efficient AI planning approach. It solves complicated planning problems in hierarchical decomposition analogous to human decision-making process. The main idea is to decompose abstract compound tasks into more specific ones until all tasks are primitive which can be executed directly. Owing to the use of specific domain knowledge, HTN planning has powerful reasoning and good support for large-scale domains [10]. In addition, it has been shown to be more expressive than classical planning representations [11]. HTN planning has been successfully applied in a great deal of real-world projects [12-14].

Several approaches have been come up with based on HTN for partially observable planning problem including contingent planning and probabilistic contingent planning. Contingent planning is characterized as uncertainty of the initial state of the world, sensing action which obtains the factual observation of belief states and the conditional plan for potential contingencies. CondSHOP2 [15] applies the conditional forward-chain search technology into

SHOP2 [16], substituting for the former forward-chain planning. It is improved to deal with uncertain states. Probabilistic contingent planning, enhanced by the presentation and the reasoning of probability, is an alternative significant approach for partial observability. C-SHOP [17], based on SHOP2 likewise, involves belief states to handle the incomplete information of the environment and sensing actions to generate a conditional plan. The improved version PC-SHOP [18] inherits all the characteristics of C-SHOP and adds the probability reasoning of action effect. But all the aforementioned works can only search a feasible solution.

Simultaneously some research are explored regarding planning based on HTN for high-quality plans. In the HTN planner SHOP2, a branch-and-bound search is used to find the plan with the minimum cost. Hogg *et al.* [7] propose an integrated approach of HTN learning with reinforcement learning to generate high-quality plans by learning the methods and their values. In [19] the messy genetic algorithm is used in HTN planning to search an optimal solution. Georgievski and Lazovik [20] present a framework of HTN planning with utility function aiming for the solution with the risk optimum. On the foundation of extending Planning Domain Definition Language 3 (PDDL3) [21], Sohrabi *et al.* [22] propose a preference-based HTN planner integrated with a branch-and-bound algorithm to find preferred plans.

The current studies of HTN planning under partial observability are unable to distinguish between the feasible plans with the various qualities, while the planning approaches based on HTN for high-quality plans concentrate upon the deterministic domains. To the best of our knowledge, HTN planning for high-quality plans under partial observability has not been addressed yet. This gap motivates us to develop a planner that can deal with that problem.

In this paper we put forward a novel probabilistic contingent planning approach based on HTN, called High-Quality Contingent Planner (HQCP), to generate high-quality plans in the partially observable environment. We extend the formalisms in HTN planning into partial observability and evaluate them regarding the cost. In addition, we explore a heuristic to guide for searching high-quality plans. Based on that, we develop the planner HQCP combining the probabilistic contingent planning with the heuristics. We empirically evaluate HQCP in two partially observable domains, Medicate and extended ZenoTravel. The experiment results show that HQCP performs effectively both in solving probabilistic contingent planning problem and in searching for high-quality plans under partial observability.

The rest of this paper is organized as follows. Section 2 presents the formalism, definition, and notation extended in this paper. In Section 3 we focus on the proposed heuristics. The integrated planning algorithm is constructed in Section 4. Finally, Section 5 surveys the empirical evaluation and the results of experiments.

## 2. Problem Definition

This section mainly discusses the necessary extensions with respect to partial observability and high quality, while the rest of the notions such as the task network are analogous to the classical HTN planning approaches [16,23]. In order to present partially observable information of environment, belief states are introduced. To observe the uncertain information, sensing actions are defined. The plan is extended as a conditional form to cope with the uncertain outcomes.

To generate high-quality plans, one necessary thing that should be taken into account first is how to quantitatively evaluate them. Since the cost can be directly evaluated in the planning domain, one natural way is to utilize the cost to distinguish different qualities [24]. Hence planning for high-quality plans turns into planning for the final goal with the minimum cost. In this section the evaluation regarding cost is discussed. The problem of probabilistic contingent HTN planning for high-quality plans is defined firstly as follows.

**Definition 1** (Problem) A probabilistic contingent HTN planning problem for high-quality plans is a 5-tuple $P = (s_0, bs, \omega_0, D, \Delta)$ where $s_0$ is the initial state, $bs$ is the belief state, $\omega_0$ is the initial task network, $D = (O, M)$ is the planning domain which is formalized by a set of operators and a set of methods, and $\Delta$ is a set of the cost of the literal $l \in s$ or $l \in bs$.

*2.1. Belief States*

**Definition 2** (Belief State): A belief state is a pair $b = (S, P)$ where $S$ is a set of ground states and $P$ is a probabilistic distribution over $s \in S$.

For presenting the incomplete information in the partially observable environment, a belief state stands for a probability distribution over a state, which refers to a set of states, instead of an exact state. Furthermore, the elements in a belief state have the same deterministic fluent and atoms, and different nondeterministic atoms and their probabilities, meaning that a belief state describes various uncertain situations over the probabilistic distribution regarding one state. Note that the sum of the probabilities of the element states should equals to 1, namely $\sum_{s \in S} b(s) = 1$. It assumes, in this study, that the real state is certain to remain with the belief states.

*2.2. Planning domain*

Like classical HTN planning, the planning domain is formalized by a pair constituted by operators and methods. The disparity that should not be omitted, however, is that the notions of operators and methods are broadened into the partially observable domain for high-quality plans.

Sensing actions aim to observe the planning environment and to obtain the observation regarding the uncertain information. In respect that sensing actions do not transform any states, they belong to actions but they should be distinguished from the other actions which change the states of objective environment [25]. Therefore, in this study we designate the actions that change the states and lack of sensing capability as actuation actions. Sensing actions lack the positive and negative effects. The accessorial portion, however, is embodied at observation. As the result of sensing, observation presents the real situation apperceived by the agent in the partially observable environment, and is treated as the precondition of the conditional plan which means that if the agent senses an observation then it executes the corresponding plan.

At the same time, the assumption of deterministic actions is too restrictive when the actuation actions fail to effect or the sensing actions may make a mistake. Therefore, we bring in the probabilistic actions to present the uncertainty in actions. Both the sensing actions and the actuation actions are evaluated by probability, which implies the probability of effecting the environment successfully by the actuation actions and the probability of making an accurate observation by the sensing actions. Consequently, as generic formalization of actions, operators are defined as below.

**Definition 3** (Operator) An operator is a pair $O = (O_{act}, O_{sense})$, $O_{act} = (name(o), pre(o), effect(o), prob(o))$, $O_{sense} = (name(o), pre(o), ob(o), prob(o))$, where $name(o)$ is the name of the operator, $pre(o)$ is the precondition, $prob(o)$ is the probability of the operator, $effect(o)$ is the positive and negative effect of the actuation operator, and $ob(o)$ is the observation of the sensing operator.

The methods defined in this study are different from the traditional HTN planning paradigms. A method has a subtask network, but a compound task can be composed by many instances of the methods.

**Definition 4** (Method) A method is a triple $M = (name(m), pre, subtask)$, where $name(m)$ is the name of the method, $pre$ is the precondition, subtask is the subtask network.

*2.3. Plan*

Instead of a set of actions, the plan is represented as the conditional form where the different branches correspond to the different observations. At the same time, each branch of the plan is evaluated by probability.

**Definition 5** (Plan) A plan is defined as $\pi = <a_1, a_2, \cdots, (O_1, \pi_{o1}, P_{o1}), (O_2, \pi_{o2}, P_{o2}), \cdots, (O_n, \pi_{on}, P_{on})>$, where $a$ is an actuation action, $O$ is the observation resulted from a sensing action, and $\pi_o$ is the corresponding plan of $O$. The conditional plan $\pi$ means that if the agent senses the observation $O$ then it carries out the plan $\pi_o$ in succession. Given the probability of the actions, the plan has the probability likewise. It is defined as $P(\pi) = \prod P(a)$

*2.4. Cost*

In this paper, the formalism of the state is represented as a set of logical atoms which are formalized as literals. For evaluating the cost of a state, we evaluate the literals firstly. As mentioned before, the states are separated into two categories, deterministic states and belief states. For deterministic states, the literals are evaluated directly. In order to evaluate the cost quantitatively, the states under particular requirements are evaluated by evaluation $\Delta$. As a part of domain knowledge, the evaluation is provided in the planning domain.

We take ZenoTravel [26] domain as example. ZenoTravel describes a problem that passengers are transported by an airplane between cities. When the airplane lands on an airport, it can be refueled by the oil supplier there. Assume that in a problem there is a state *s=(supplier B unoccupied)*. The cost of the state is $\Delta(s) = 100$, which represents that when supplier B is not occupied refueling at B costs 100.

For a belief state, the evaluation equals with the expectation of the cost of the state under the probability distribution. Although there are the sensing actions that are capable of sensing the real environment to acquire the deterministic state, the evaluation of the belief state happens before the execution of the sensing action. Hence the cost of the belief state is evaluated by expectation.

For example, we extend ZenoTravel to be a partially observable domain. Refueling situation is uncertain. Specifically when the airplane lands on the airport but the oil supplier is occupied by other planes, it cannot be refueled. The occupation, however, cannot be informed in advance. Thereby belief state *b=(((supplier B occupied) 0.1) ((supplier B unoccupied) 0.9))* represents supplier B is occupied with the possibility 0.1 and unoccupied with 0.9. As the defined above, belief state *b* has cost of 130.

Some works for high-quality plans in HTN planning [16,20] and in classic planning [4] assign the operators the attribute of cost, which means each operator has the cost and its constant cost is given in the planning domain. This assumption, however, has limitation. Since the operators are an abstract form, different actions instantiated from one operator have different costs. For instance, operator *!fly* describes an airplane flies from one city to the other city. Its cost is considered as the consumption of the fuel during the voyage. Hence the longer the flight route is, the more fuel is consumed. This motivates us to evaluate the cost of actions instead of operators.

As a general formalization, an operator is noted by variables and constraints, and only when it is instantiated into actions, it has the real sense. Based on the definition of the cost above, the cost of the action is assigned by the cost of the ground literals in the precondition. Formally $\Delta(a) = \sum \Delta(l)$ *where* $l \in pre(a)$.

The definition of the task network in this study is analogous to the classical HTN planning approaches. But the tasks in the task network should be evaluated by the cost. The cost of any uninstantiated task is 0, no matter primitive tasks or compound task. The cost of the tasks are updated as the tasks are instantiated. The procedure of updating is elaborated in the following sections.

A method defines a way by which a compound task is decomposed into the subtasks. The cost of the method is defined as $\Delta(m) = \sum \Delta(l)$ where $l \in pre(m)$. The plan in this study consists of a set of actions in a conditional form, and the cost of the plan is determined by the actions. Therefore, the cost of the plan is defined as $\Delta(\pi) = \sum \Delta(a)$ where $a \in \pi$.

## 3. Heuristics

In this section, we discuss the heuristics for high-quality plans. In HTN planning, the task network, as control strategy, guides the reasoning of actions [1]. Planning goal is replaced by the initial task network in HTN planning problem. It means that if there exists a plan the initial task network is certain to be decomposed accurately to achieve the final goal via the given planning domain.

As defined above, the cost of any uninstantiated task is 0, no matter primitive tasks or compound task. The cost of the tasks are updated as the tasks are instantiated and the planning goes forward. After a primitive is instantiated into an action, the cost of the primitive task is assigned by the cost of the action. When a compound task is instantiated by a method, the cost of the compound task is updated by the cost of the method. Among the feasible plans that are decomposed through the compound task $t$, $\Delta^*(t)$ denotes the cost of the plan with the minimum cost. If there is no plan that can be decomposed through $t$, $\Delta(t) = \infty$. We define a heuristic function $\Delta(t)$ which is given as Formula 1. c and p represent the compound task and the primitive task in $subtask(t)$, respectively. $subtask(t)$ is the subtask set. The formula is calculated, and $\Delta(t)$ is assigned by the cost of $subtask(t)$ that have the minimum cost.

$$\Delta(t) = \min\{\sum_{c,p \in subtask(t)} (\Delta(c) + \Delta(p))\} \tag{1}$$

At the same time, we define Delta($t$) as the algorithm that calculates the heuristic value $\Delta(t)$ in HTN planning. The detailed procedure is presented in Algorithm 1.

We briefly discuss the admissibility of the heuristic algorithm. Assume that the algorithm ends at a suboptimal plan n′, namely $\Delta(n') > \Delta^*(n)$. There is a plan that belongs to the optimal plans and that is $\Delta(n_0) \leq \Delta^*(n) < \Delta(n')$. Therefore $n_0$ is expanded prior to n′ and it is not possible to be expanded. That contradicts with the assumption above. For this reason the algorithm is admissible under the circumstance that $\Delta(t) \leq \Delta^*(t)$.

---

**Algorithm 1** Delta(t)

---
1. **begin**
2.     c is the compound task and p is a primitive task, $c, p \in subtask(t)$
3.     $\Delta(t) \leftarrow \min\{\sum (\Delta(c) + \Delta(p))\}$
4.     **return** $\Delta(t)$
5. **end begin**

---

## 4. Planning Algorithm

The study aims to plan in the partially observable environment for high-quality plans. Therefore it requires combining partial observability with optimizing procedure. More specifically, the obtained plan is capable not only to deal with partial observability analyzed before but also to be a high-quality one at the evaluation of some objectives. In this circumstance, we explicate the planning algorithm that integrates the heuristics with probabilistic contingent planning. In this section, we present the planning algorithm based on the heuristics above. Section 4.1 gives an overview of the planning framework. The detailed description for the main steps will be given in Section 4.2.

*4.1. Algorithm framework*

The algorithm framework is chiefly organized in four steps: instantiate the task, update the cost, backtrack or plan forward. If each task in the current task network is the one that minimizes it, the task network is consistent. If there exists a task q that cannot minimize the task network, the task network is not consistent at q.

Step 1: instantiate the task
The current task, a primitive task or a compound task, is instantiated by an available instance with the minimum cost. The cost of the current task is assigned at the same time.

Step 2: update the cost
The cost of the tasks in the current task network are updated from the bottom up by Delta($t$). In the meantime each task will be checked to verify the consistency as soon as being updated. If a task is not consistent, the whole updating process will be suspended. Only when the current task network is consistent, planning goes forward.

Step 3: backtrack

Backtracking and planning forward are mutually exclusive. If a task is not consistent, backtracking will be executed. Backtracking includes removing of the actions from the plan, reverting of the states, withdrawing the task network.

Step 4: plan forward
When the task network is consistent, the planning goes forward. At the same time the next iteration is called in recursion.

*4.2. Planning algorithm*

The planner takes the planning problem $P = (s_0, bs, \omega_0, D, \Delta)$ as the input of the planner, which includes the initial state $s_o$, the belief state $bs$, the initial task network $\omega_0$, the planning domain $D$ and the evaluation of cost. Finally the planning algorithm aims to generate a plan for $P$. The planning procedure is shown as Algorithm 2.

---

**Algorithm 2** $HQCP(s_0, bs, \omega_0, D, \Delta)$

**Initialize:** $\pi \leftarrow \emptyset, s \leftarrow s_0, \omega \leftarrow \omega_0, D = \{O, M\}, \forall t, \Delta(t) \leftarrow 0$

1.    **begin**
2.      **if** $\omega = \emptyset$ **then return** $\pi$
3.      pop any $t \in \omega$
4.      **if** $t$ is an actuation primitive task **then**
5.        $Active \leftarrow \{(a, \sigma) \mid a$ is a ground instance of the operator o, $\exists \sigma, \sigma(t) = head(a), s \models pre(a)\}$
6.        **while** $A \neq \emptyset$ **do**
7.          select a pair $(a, \sigma) \in Active$ by min $\Delta(a)$
8.          update $(t)$
9.          **if** $\omega$ is consistent **then**
10.            $s' \leftarrow (s - effect^-(a)) \cup effect^+(a)$
11.            $\omega' \leftarrow \omega - t$
12.            $\pi \leftarrow a \cup HQCP(s', bs, \omega', D, \Delta)$
13.          **if** $\pi$ **is not** failure **then return** $\pi$
14.        **end while**
15.        **if** $A = \emptyset$ **then return** failure
16.      **else if** $t$ is a compound task **then**
17.        $Active \leftarrow \{(m, \sigma) \mid m$ is an instance of the method M, $\exists \sigma, \sigma(t) = head(m), s \models pre(m)\}$
18.        **while** $Active \neq \emptyset$ **do**
19.          select a pair $(m, \sigma) \in Active$ by min $\Delta(m)$
20.          update $(t)$
21.          **if** $\omega$ is consistent **then**
22.            $\omega' \leftarrow (\omega - u) \cup subtask$
23.            $\pi \leftarrow HQCP(s, bs, \omega', D, \Delta)$
24.          **if** $\pi$ **is not** failure **then return** $\pi$
25.        **end while**
26.        **if** $C = \emptyset$ **then return** failure
27.      **else if** $t$ is a sensing primitive task **then**
28.        **while** $bs \neq \emptyset$ **do**
29.          pop $s_o \in bs$
30.          $\pi_o \leftarrow Observe(s_o)$
31.          $s' \leftarrow s \cup s_o$
32.          $\pi' \leftarrow (\pi_o \cdot HQCP(s', bs, \omega, D, \Delta))$
33.          $\pi \leftarrow \pi \cup \pi'$
34.        **end while**
35.        **if** $bs = \emptyset$ **then return** failure

| | |
|---|---|
| 36. | **if** $\pi$ **is not** failure **then return** $\pi$ |
| 37. | **end begin** |

The algorithm initializes the indispensable variables at the beginning. Plan $\pi$ is assigned by an empty set. State $s$ and task network $\omega$ take $s_0$ and $\omega_0$ as the initial value respectively. As defined, planning domain $D$ consists of the operators and the methods.

The framework of the algorithm can be regarded as a loop that plans each step and moves forward recursively. When the plan is generated successfully or the algorithm returns failure, the whole cyclic procedure stops and the planning ends.

The tasks in the task network are defined by three categories: compound tasks, actuation primitive tasks and sensing primitive tasks, corresponding to methods, actuation operators and sensing operators respectively. Hence the algorithm proceeds dissimilarly in the light of the three ways. From line 4 to line 15, when the current task is an actuation primitive task, an actuation operator $o$ is instantiated by a substitution $(a, \sigma)$ at state s that satisfies pre($a$). Line 6 checks actions A. When A is not null, action $a$ is instantiated, state s is updated by $(s - effect^-(a)) \cup effect^+(a)$, the task network is updated and plan $\pi$ is appended with $a$.

Line 16 to line 26 describes the situation where the current task is a compound task. C is a set that consists of the instances by decomposing the current task under the substitution σ. If C is not null, the instance that minimizes the cost is chosen. All the tasks in the task network will be updated from the bottom up at line 20. The detailed procedure of updating will be discussed in the following section. If the task network is checked to be not consistent at a task $q$, updating will be suspended and backtracking happens to search other available instances of the methods. Otherwise the task network is renewed at line 22 and the planning proceeds.

When the current task is a sensing primitive task, the algorithm senses the whole distribution of the belief state for uncertainty respectively, and calls HQCP recursively, which forms a conditional plan. From line 27 to line 36, the current task is a sensing primitive task. The algorithm senses every atom in the belief state, and the observation is added into the corresponding plan. At line 32, the algorithm is invoked recursively to form the conditional plan.

| | **Algorithm 3** update ($t$) |
|---|---|
| 1. | **begin** |
| 2. | $t \in$ subtask($\bar{t}$), $t \in \omega$, $\bar{t} \in \omega$ |
| 3. | $\Delta(\bar{t}) \leftarrow$ Delta($\bar{t}$) |
| 4. | **if** $t$ is not consistent **then** |
| 5. | **return** failure |
| 6. | **else if** $\bar{t} = \emptyset$ |
| 7. | **return** true |
| 8. | **else** |
| 9. | update ($\bar{t}$) |
| 10. | **end begin** |

Algorithm 3 describes the detailed procedure of updating the cost of all the tasks in the task network. As the heuristics, the cost of the father task $\bar{t}$ is updated at line 3. This procedure repeats recursively until a task is not consistent or the whole task network is consistent

## 5. Empirical Study

We evaluate HQCP's performance in two planning domains: medicate domain and extended ZenoTravel domain. Medicate domain is a typically partially observable domain, in which we compare HQCP with the state-of-the-art HTN-based probabilistic contingent planner PC-SHOP. Next, we extend the standard ZenoTravel into a partially observable domain in which the planner evaluates the probability and the cost. Section 1 aims to verify the performance of dealing with partial observability. Section 2 attempts to show the performance of planning for high-quality plans in the partially observable environment. All the experiments are run on a Windows 7 machine with 4 GHz Pentium 4 and 4GB of RAM. Common LISP is chosen as the programming language.

*5.1. Medicate domain*

The medicate domain [2] takes consideration of a patient who may have n possible infections or may be healthy, but we do not know which infection the patient has. The treatment for the patient involves diagnosing the infection and accordingly taking a proper medicate action to cure it. Inaccurate remedy, however, will kill the patient. In the planning domain, there is one sensing operator that diagnoses the disease and one actuation operator that medicates the patient using the corresponding remedy. The planning goal is to cure the patient. The plan outputted from the planner diagnoses each disease and cures it with right medicate action conditionally.

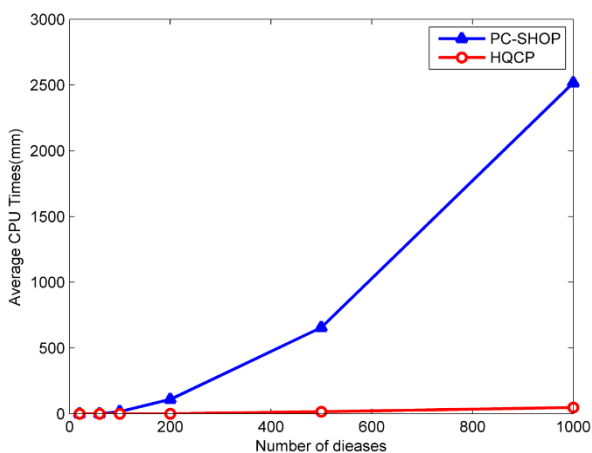

Fig 1. Average runtime for the medicate domain with n diseases.

Fig. 1 shows the runtime of the two planners. Every experiment is run five times and the average CPU time is recorded. Fig. 1 indicates that HQCP is faster than the contrastive planner. With the difficulty of the planning problem increasing and the scale of the state space expanding, the efficiency of HQCP does not decline dramatically while the contrastive planner cost much time. The experiment results also show that HQCP is capable to cope with real engineering problems with a high complexity. In our opinion, the better performance demonstrated by Fig. 1 comes from the more succinct algorithm structure and the more efficient heuristic rules.

*5.2. Extended ZenoTravel domain*

In order to illustrate the performance of HQCP for high-quality plans in the partially observable environment, we extend standard ZenoTravel, one of the standard domains in the planning competition of the International Conference on AI Planning and Scheduling (AIPS-2002) [27], into the partially observable domain. ZenoTravel is extended to verify the capabilities of the planning for high-quality plans. As the expressive power of the numeric tracks is used, searching for high-quality plans is much harder [5]. In the ZenoTravel domain, an airplane transports people from city A to city C transiting at city B. The airplane has two navigation modes, zoom and fly. Zoom, the fast movement, flies faster but consumes more fuel. The airplane needs to refuel when it lands on an airport if the zoom mode is opened. Passengers board and debark from the airplane at the airport. Therefore, there are five actions: board, debark, fly, zoom and refuel. Since the temporal problem is inevitable, all five actions have the execution durations, and refueling can be carried simultaneously with boarding or debarking.

One key issue in the extended ZenoTravel domain is the numeric cost. The domain requires to obtain the plan with the minimum fuel consumption. Furthermore, the airplane needs to refuel when it lands on an airport, while the refueling situation is uncertain. Specifically when the airplane lands on the airport but the oil supplier is occupied by the other planes, the airplane cannot be refueled. The occupation, however, cannot be informed in advance.

Table 1

The plans of the extended ZenoTravel problems

| Plan 1 | | | Plan 2 | | |
|---|---|---|---|---|---|
| actions | start-time | end-time | actions | start-time | end-time |
| (!Observe usable A) | | | (!Observe usable A) | | |
| (!board-passenger 20) | 16:35 | 16:55 | (!refuel-at A) | 16:35 | 16:55 |
| (!fly A B) | 16:55 | 19:10 | (!board-passenger 20) | 16:35 | 16:55 |
| (!debark-passenger 10) | 19:10 | 19:25 | (!zoom A B) | 16:55 | 18:30 |
| (!board-passenger 30) | 19:10 | 19:25 | (!debark-passenger 10) | 18:30 | 18:45 |
| (!fly B C) | 19:25 | 21:20 | (!board-passenger 30) | 18:30 | 18:45 |
| (!debark-passenger 40) | 21:20 | 21:35 | (!fly B C) | 18:45 | 20:40 |
| (!Observe unusable A) | | | (!debark-passenger 40) | 20:40 | 20:55 |
| (!board-passenger 20) | 16:35 | 16:55 | (!Observe unusable B) | | |
| (!fly A B) | 16:55 | 19:10 | (!board-passenger 20) | 16:35 | 16:55 |
| (!debark-passenger 10) | 19:10 | 19:25 | (!fly A B) | 16:55 | 19:10 |
| (!board-passenger 30) | 19:10 | 19:25 | (!debark-passenger 10) | 19:10 | 19:25 |
| (!fly B C) | 19:25 | 21:20 | (!board-passenger 30) | 19:10 | 19:25 |
| (!debark-passenger 40) | 21:20 | 21:35 | (!Observe usable B) | | |
| | | | (!refuel-at B) | 19:10 | 19:25 |
| | | | (!zoom B C) | 19:25 | 20:55 |
| | | | (!board-passenger 40) | 20:55 | 21:10 |
| | | | (!Observe unusable B) | | |
| | | | NULL | | |

Table 1 shows the different plans under the different deadlines. In Plan 1, the deadline of arriving city C is 21:30. In order to cut the cost, the airplane uses fly mode. However, in Plan 2 the deadline of arriving city C is 21:00. The airplane opens fly and zoom modes. The plans observe whether the oil supplier is usable when the airplane lands on an airport. As a result of assistance of the heuristics, the two plans are both the plans with the minimum costs under the temporal requirements.

## 6. Conclusions

In this paper we have proposed a probabilistic contingent planner based on HTN for high-quality plans under partial observability. We have extended the formalisms in HTN planning into partial observability and evaluated them regarding the cost. Next, we have developed the heuristics for searching high-quality plans. A complete planning algorithm has been constructed to describe the whole procedure of planning. Finally, two partially observable domains have been selected for verifying the performance of the planning approach. The first domain has illustrated that the planer has efficiency in coping with partial observability. The second one has shown its capability of high-quality planning in the partially observable environment.